\documentclass[10pt,twocolumn,letterpaper]{article}

\usepackage{cvpr}              

\usepackage{graphicx}
\usepackage{amsmath}
\usepackage{amssymb}
\usepackage{booktabs}

\usepackage{multirow}
\newcommand{\myparagraph}[1]{\vspace{3pt}\noindent{\bf #1}}

\usepackage[pagebackref,breaklinks,colorlinks]{hyperref}

\usepackage[capitalize]{cleveref}
\crefname{section}{Sec.}{Secs.}
\Crefname{section}{Section}{Sections}
\Crefname{table}{Table}{Tables}
\crefname{table}{Tab.}{Tabs.}

\def\cvprPaperID{*****} 
\def\confName{CVPR}
\def\confYear{2023}

\begin{document}

\title{Resolving Class Imbalance Problem for LiDAR-based Object Detector\\by Balanced Gradients and Contextual Ground Truth Sampling}

\author{Daeun Lee$^1$, Jongwon Park$^2$, and Jinkyu Kim$^{1,*}$\\
$^1$Department of Computer Science and Engineering, Korea University\\
$^2$Autonomous Driving Center, Hyundai Motor Company R\&D Division\\
{$^*$Correspondence: \tt\small jinkyukim@korea.ac.kr}
}
\maketitle

\begin{abstract}
    An autonomous driving system requires a 3D object detector, which must perceive all present road agents reliably to navigate an environment safely. However, real-world driving datasets often suffer from the problem of data imbalance, which causes difficulties in training a model that works well across all classes, resulting in an undesired imbalanced sub-optimal performance. In this work, we propose a method to address this data imbalance problem. Our method consists of two main components: (i) a LiDAR-based 3D object detector with per-class multiple detection heads where losses from each head are modified by dynamic weight average to be balanced. (ii) Contextual ground truth (GT) sampling, where we improve conventional GT sampling techniques by leveraging semantic information to augment point cloud with sampled ground truth GT objects. Our experiment with KITTI and nuScenes datasets confirms our proposed method's effectiveness in dealing with the data imbalance problem, producing better detection accuracy compared to existing approaches.
\end{abstract}
\vspace{-1em}

\section{Introduction}
LiDAR-based detectors have been widely adopted in the autonomous driving system for capturing 3D scene perception and understanding. Such an autonomous driving system must detect all possible other road agents (or objects) to navigate an environment safely. Thus, a reliable LiDAR-based detector requires dealing equally with different road agents (or objects), e.g., cars, cyclists, barriers, or construction vehicles.

\begin{figure}[t]
    \begin{center}
        \includegraphics[width=\linewidth]{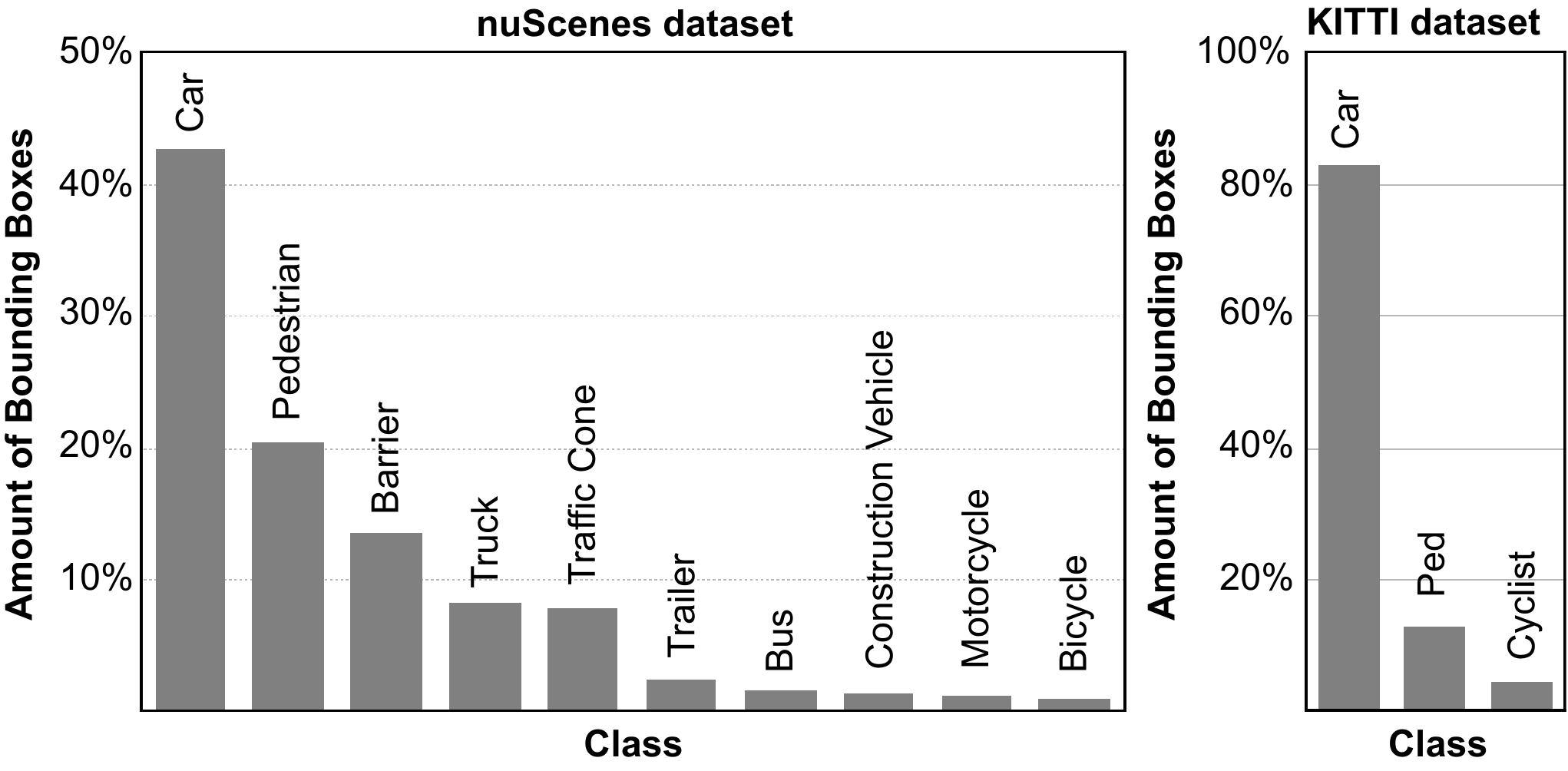}
    \end{center}
    \caption{Class distributions for two 3D object detection datasets: nuScenes~\cite{caesar2020nuscenes} (left) and KITTI~\cite{geiger2013vision} (right).}
    \label{fig:freq}\vspace{-1em}
\end{figure}

However, real-world driving datasets (e.g., KITTI and nuScenes) suffer from the problem of imbalance where a dataset contains unequal (or severely skewed) class distribution. As shown in Figure~\ref{fig:freq}, objects such as cars (42.63\%) have a higher percentage compared to the percentage of other classes, such as bicycles (1.03\%), motorcycles (1.11\%), or construction vehicles (1.39\%). Similarly, in the KITTI dataset, cars (82.99\%) have the majority of instances, while pedestrians (12.76\%) or cyclists (4.24\%) are underrepresented. Such data imbalance would cause difficulties in training a 3D object detector that reliably works well across all different classes, resulting in an undesired imbalanced quality.

Multi-task learning techniques have been applied to address this data imbalance problem by viewing multi-class joint detection as multi-task learning. In this work, we explore applying such multi-task learning techniques to address the data imbalance problem in the LiDAR-based 3D object detection task. Specifically, we focus on answering two key questions: (i) constructing multi-task network architecture and (ii) balancing feature sharing across different tasks. For (i), we use per-class multiple detection heads instead of a single head. Each detection head is encouraged to learn class-specific features while sharing a backbone, which is trained to extract universal features. For (ii), we explore applying existing multi-task loss balancing techniques to improve the overall performance of different detection heads. Specifically, we apply Dynamic Weight Average (DWA, \cite{liu2019end}) that tunes gradients for different object categories based on the rate of loss changes for each head to learn average task weighting over time. We empirically observe that combining multi-headed architecture and gradient balancing techniques significantly improves detection accuracy. 

Another story is data augmentation, which can make class distribution smoother by making the model sees rare classes more often during training. Conventionally, ground truth (GT) sampling has been widely used. GT sampling collects all ground truth points inside the labeled bounding box into a database, and some of them are randomly introduced to the current training frame via concatenation. However, this does not consider where to place these objects. We, in fact, observe ground truth points are often introduced in a random position where that object is rarely observed in the real world. Thus, we propose {\it contextual} GT sampling that leverages semantic scene information to present ground truth points in a more natural position, e.g., a sidewalk for pedestrians. Our experiment shows that our contextual GT sampling provides extra performance gain, especially for minor classes. 

Our approach is mostly close to Zhu~\etal~\cite{zhu2019class} (CBGS) in that they also use multiple detection heads and data augmentation techniques, i.e., GT sampling~\cite{yan2018second}. However, our work differs from it as follows: (i) we explore using multi-task learning techniques, including multiple detection heads with loss balancing techniques, to improve overall detection performance across all categories. CBGS focused on utilizing multi-headed architecture with a uniform scaling, which minimizes a uniformly weighted sum and does not consider dynamically modifying weights like ours. (ii) we propose contextual GT sampling, which addresses issues with conventional GT sampling and results in better detection accuracy.

We summarize our contributions as follows:
\begin{itemize}
    \vspace{-.5em}
    \item Inspired by multi-task learning, we propose a multi-headed LiDAR-based 3D object detector where losses for each head are balanced by dynamic weight average (DWA).
    \item Combined with multi-headed architecture, we propose contextual ground truth sampling, which improves conventional ground truth (GT) sampling by leveraging semantic scene information to introduce GT objects in a more realistic position.
    \item We conduct various experiments to demonstrate the effectiveness of our proposed approach with widely-used public datasets: KITTI and nuScenes. Our experiments show that multi-task learning techniques combined with our contextual GT sampling significantly improve the overall detection performance, especially for minor classes.
\end{itemize}

\begin{figure*}[t]
    \begin{center}
       \includegraphics[width=\linewidth]{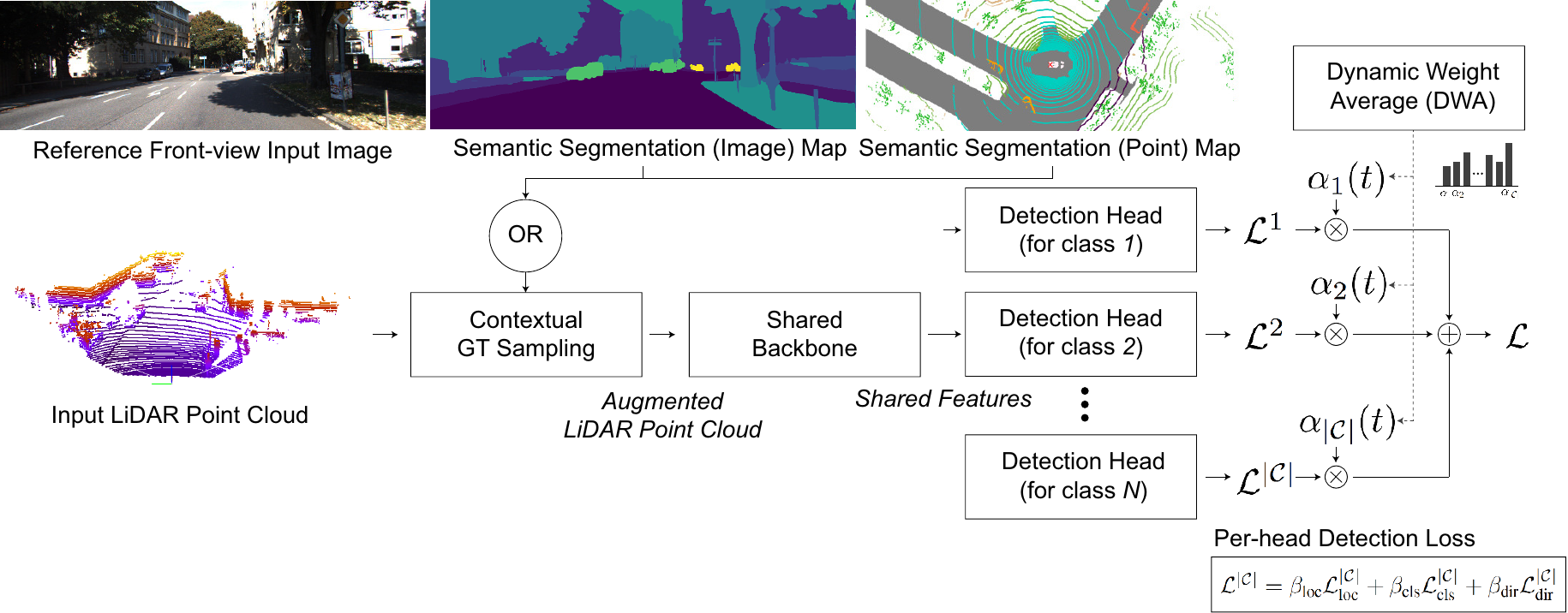}
    \end{center}
    \caption{An overview of our proposed method to address the data imbalance problem in the LiDAR-based 3D object detection task. Our model consists of two main parts: (1) Per-class multi-headed architecture where detection losses $\mathcal{L}^{c}$ for $c\in\mathcal{C}$ for each head are balanced by dynamic weight average (DWA). (2) Contextual ground truth (GT) sampling, which is built upon conventional GT sampling and improves it by leveraging semantic scene information (either semantic segmentation image map or semantic point map) to place ground truth points in a more realistic position.}
    \label{fig:multi-headed}\vspace{-1em}
\end{figure*}

\section{Related Work}
\subsection{3D Object Detection}
A landmark work in the LiDAR-based 3D object detection is VoxelNet~\cite{zhou2018voxelnet}, an end-to-end trainable model that first voxelized a point cloud, and each equally spaced voxel is encoded as a descriptive volumetric representation. Given these features, conventional 2D convolutions are used to generate and regress its region proposals. Yan~\etal~\cite{yan2018second} used sparse 3D convolutions to accelerate heavy computations of earlier LiDAR-based works. PointPillars~\cite{lang2019pointpillars} is another landmark work that speeds up the encoding of 3D volumetric representation by dividing the 3D space into pillars (instead of voxels). A more sophisticated architecture is also used to achieve better detection results. PointRCNN~\cite{Shi_2019_CVPR} used a two-stage architecture to refine the initial 3D bounding box proposals. Part-A2~\cite{shi2020points} focuses on leveraging intra-object parts for better results. PV-RCNN~\cite{shi2020pv} and PV-RCNN++~\cite{shi2021pv} simultaneously process coarse-grained voxels and the raw point cloud. Recently, CenterPoint~\cite{yin2021center} applied a key-point detector that predicts the geometric center of objects. Similarly, Voxel RCNN~\cite{deng2021voxel} used coarse voxel granularity to reduce the computation cost, retaining the overall detection performance. In this work, we focus on improving data imbalance problems in LiDAR-based object detection. Thus, we do not claim a novel 3D object detector; rather, we rely on existing landmark work PointPillars~\cite{lang2019pointpillars}, PV-RCNN~\cite{shi2020pv}, and Voxel RCNN~\cite{deng2021voxel} to demonstrate the effectiveness of our proposed approach. Note that, ideally, our approach is applicable to others as well.

\myparagraph{Lidar Points Augmentations.}
Data augmentation has been widely applied to LiDAR-based 3D object detection for various reasons: (i) improving point cloud quality by upsampling a low-density point cloud~\cite{yu2018pu, yifan2019patch} or by point cloud completion for occluded regions~\cite{yuan2018pcn, yang2018foldingnet, chen2019unpaired, xie2020grnet}. (ii) Improving the robustness of object detection by global and local augmentations. Choi~\etal~\cite{choi2021part} randomly augmented sub-partitions of GT objects (e.g., dropping points in a certain sub-partition)~\cite{choi2021part}. Zheng~\etal~\cite{zheng2021se} divided each ground truth object into six (inward facing) pyramids, then augmented them with random dropout, swap, and sparsifying operations. (iii) Improving generalization power by augmenting clear weather point clouds with adverse conditions via physical modelings, such as fog~\cite{hahner2021fog} or snowfall~\cite{hahner2022lidar}. (iv) Augmenting LiDAR-based features with other modalities, such as images~\cite{wang2021pointaugmenting, vora2020pointpainting}. (v) Smoothing class distribution by sampling ground truth objects from the (offline) database and introducing them to the current scene (GT sampling, \cite{yan2018second}). In this work, similar to (v), we focus on smoothing the density of each class to address the data imbalance problem (i.e., improving the detection accuracy of rare objects while maintaining that of common objects). Thus, we start with GT sampling~\cite{yan2018second} as a baseline.

\subsection{Gradient Balancing}
Multitask learning has been widely explored to share features across different tasks while making task-specific multiple predictions, such as for multi-domain image classification~\cite{rebuffi2017learning}, post estimation and action recognition~\cite{gkioxari2014r}, or depth estimation and semantic segmentation~\cite{eigen2015predicting, misra2016cross}. Key questions in multitask learning are (i) constructing multitask network architectures and (ii) balancing feature sharing across different tasks. To address the latter, studies reported that multitask loss balancing techniques improve the overall performance for different tasks~\cite{misra2016cross, kendall2018multi}. Kendall~\etal~\cite{kendall2018multi} modified the loss function based on task uncertainty, and GradNorm~\cite{chen2018gradnorm} dynamically tuned gradient magnitudes and showed it improves accuracy and reduces overfitting across different tasks. Dynamic Task Prioritisation~\cite{guo2018dynamic} prioritized difficult tasks based on performance metrics. Dynamic Weight Average (DWA, \cite{liu2019end}) is also proposed to use the rate of loss changes for each task to learn average task weighting over time. In this work, we explore applying recent multitask learning techniques to tune gradients for different object categories and reduce the data imbalance problem.

\section{Method}
\subsection{Per-class Detection Heads with Gradient Balancing}

\myparagraph{Multi-heads Architecture.}
We first adopt a multi-task learning (MTL) strategy, which aims to learn multiple tasks jointly by leveraging the shared knowledge of all the tasks at hand. MTL is effective in reducing the data sparsity problem where the number of labeled examples for each task is insufficient to optimize a model. This is because MTL can aggregate all labeled data and utilize more data from different tasks to obtain a more accurate learner with generalizable representations for multiple tasks. As reported in existing works~\cite{zhu2019class}, MTL is also effective in improving the performance of multi-category joint detectors. 

As shown in Figure~\ref{fig:multi-headed}, we apply the above-mentioned multi-task learning strategy by utilizing multiple object detection heads for each category with a shared encoder (i.e., backbone). This multi-headed architecture prevents the risk of overfitting in a particular dominant task (e.g., a single-head model would more overfit in detecting common objects than the rare). It also reduces the data imbalance problem when combined with data augmentation techniques (which we will explain in Section~\ref{sec:context-aware-augmentation}).

\myparagraph{Point Cloud Encoder.}
Our model is built upon a seminar work, PointPillars~\cite{lang2019pointpillars}, though our model is easily applicable to other LiDAR-based 3D object detectors. Following PointPillars, we encode a point set $\mathcal{P}$ into an evenly-spaced grid of $M\times N$ pillars with x-y coordinates. Points in each pillars are augmented with a tuple ($x_c$, $y_c$, $z_c$, $x_p$, $y_p$), where $x_c$, $y_c$, and $z_c$ are distances to the mean of all points in the pillar, and $x_p$ and $y_p$ are offsets from the pillar center. We then apply a simplified PointNet~\cite{qi2017pointnet} architecture to encode each point and aggregate features into a single feature vector per pillar by a max operation. The resulting $M\times N$ feature map is then processed through a backbone, reshaping it to $W\times H$. Detection heads then share this resulting feature map for the final verdict.

\myparagraph{Detection Heads.}
Following PointPillars~\cite{lang2019pointpillars}, we use the Single Shot Detector (SSD) setup as the detection head, matching predictions to the ground truth using 2D Intersection of Union (IoU). Each detection head is trained with the following three types of losses: (i) localization loss $\mathcal{L}_{\textnormal{loc}}$, (ii) object classification loss $\mathcal{L}_{\textnormal{cls}}$, and (iii) heading loss $\mathcal{L}_{\textnormal{dir}}$. The total loss is as follows: 

\begin{equation}     
    \mathcal{L} = \frac{1}{N_{\textnormal{pos}}}\sum_{c\in\mathcal{C}}\alpha_c(t) (\beta_{\textnormal{loc}}\mathcal{L}_{\textnormal{loc}}^{c} + \beta_{\textnormal{cls}}\mathcal{L}_{\textnormal{cls}}^{c} + \beta_{\textnormal{dir}}\mathcal{L}_{\textnormal{dir}}^{c})
\end{equation}

we accumulate all losses from $|\mathcal{C}|$ detection heads where $\mathcal{C}$ is a set of categories. The number of positive anchors is denoted by $N_{\textnormal{pos}}$ and hyperparameters $\beta_{\textnormal{loc}}$, $\beta_{\textnormal{loc}}$, and $\beta_{\textnormal{loc}}$ are set by default as 2, 1, and 0.2, respectively. Note that we use $\alpha_c(t)$ as a loss weight determined at each timestep $t$ to tune the loss and fix data imbalances in gradient norms, which we will explain in detail in the next section. In Figure~\ref{fig:multi-headed}, we describe our LiDAR-based object detector with per-class detection heads with gradient balancing. 

\myparagraph{Balancing Gradients.}
To determine the balancing weight $\alpha_c(t)$ at each timestep $t$ for a detection head $c$, we use a technique called dynamic weight average (DWA, \cite{liu2019end}). We use DWA to compute weights for each head $c$ as follows:
\begin{equation}
    \alpha_c(t) = \frac{|\mathcal{C}|\textnormal{exp}(w_c(t-1)/T)}{\sum_c \textnormal{exp}(w_c(t-1)/T)}
\end{equation}
where $w_c(t)$ is defined as the relative descending rate at timestep (or iteration) $t$ and defined as follows: $w_c(t-1)=\mathcal{L}_c(t-1)/\mathcal{L}_c(t-2)$ where $\mathcal{L}_c(t)$ is the averaged loss value from a detection head $c$. We use a temperature $T$ to control the strength of gradient balancing. As reported in \cite{liu2019end}, we use the loss $\mathcal{L}_c(t)$ averaged over several iterations to reduce unstable training due to uncertainties from stochastic gradient descent and random training data selection.

\subsection{Context-aware LiDAR Points Augmentation}\label{sec:context-aware-augmentation}
Data imbalance problem is often observed in various perception datasets for autonomous driving. Common objects, such as cars, generally are more numerous (by a large margin) than rare object classes, such as construction vehicles or traffic cones. Such data imbalance significantly limits balancing the network's final performance per class. One way to solve this problem is via a data augmentation approach, which can make class distribution smoother by making the model see rare classes more often during training. 

\myparagraph{Sampling Ground Truths from Database (GT Sampling).}
A common practice in LiDAR-based data augmentation is sampling ground truths from the database called GT sampling~\cite{yan2018second}. All ground truth points inside the labeled bounding box (with their labels) are collected in an offline database. Some ground truth points are randomly chosen from this database during training and placed into the current frame of point clouds via concatenation, simulating objects from different frames or environments. Thus, the average density of rare classes can be improved.

\myparagraph{Contextual GT Sampling.}
Though a simple filtering rule is used to ensure sampled objects do not collide with other objects, it does not consider {\it where} to place these objects. For example, as shown in Figure~\ref{fig:contextual_sampling}, ground truth points of pedestrians are introduced to a random position where pedestrians are rarely observed (e.g., on the road intended for vehicles). To address this, we advocate using prior semantic information to introduce objects in a more realistic position (e.g., on a sidewalk for pedestrians).

{
 \setlength{\tabcolsep}{4pt}
 \renewcommand{\arraystretch}{1.3} 
\begin{table}[t]
	\begin{center}
	\caption{Object categories and their associated semantic labels (based on KITTI~\cite{geiger2013vision} and nuScenes~\cite{caesar2020nuscenes} dataset) for contextual GT sampling. {\it Abbr.} C.V.: Construction Vehicle, T.C.: Traffic Cone.}
	\label{tab:contextual_association}
    	\resizebox{\linewidth}{!}{%
    	\begin{tabular}{@{}lll@{}} \toprule
    	    Dataset & Object Category & Associated Semantic Labels \\\midrule
    	    \multirow{3}{*}{NuScenes~\cite{caesar2020nuscenes}} & Pedestrians & Sidewalk\\
    	    & Car, Truck, Bus, Trailer, C.V. & Drivable Surface\\
    	    & Motorcycle, Bicycle, Barrier, T.C. & Sidewalk, Drivable Surface\\\midrule
    	    \multirow{3}{*}{KITTI~\cite{geiger2013vision}}& Pedestrian & Sidewalk\\
    	    & Car & road \\
    	    & Cyclist & sidewalk, road\\
    	    \bottomrule
        \end{tabular}}
     \end{center} \vspace{-1em}
\end{table}
}

\begin{figure}[t]
    \begin{center}
       \includegraphics[width=\linewidth]{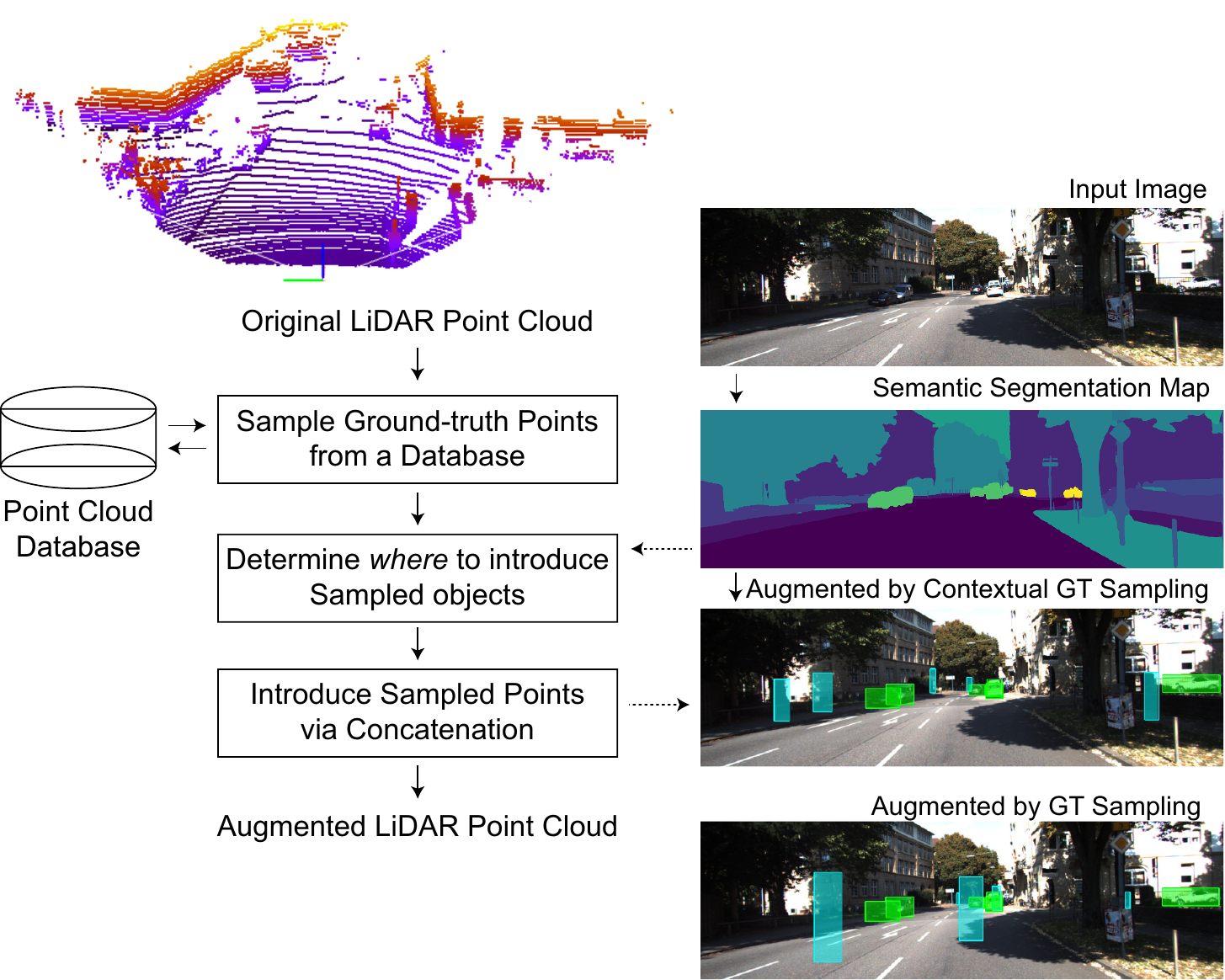}
    \end{center}
    \caption{An overview of our proposed contextual GT Sampling. We first sample some of the ground truth LiDAR points from a database. Based on a semantic segmentation map, we then identify potential regions where objects can be plausibly observed (e.g., pedestrians on the sidewalk). Conventional GT sampling does not consider this.}
    \label{fig:contextual_association}
\end{figure}

Formerly, given a 2D semantic segmentation map, we first identify potential image regions where objects are plausibly or commonly observed. In Table~\ref{tab:contextual_association}, we summarize objects and their associated semantic labels (to be placed). Based on a given camera's known intrinsic and extrinsic parameters, the 3D geometric center points of a given object bounding box are first projected into the ground (set $z$ as 0) and then projected into a 2D image plane to determine semantic information of the region. Finally, objects introduced to non-associated regions (e.g., pedestrians on a roadway) are removed by a filtering criterion. We explain an overview of our contextual GT sampling process in Figure~\ref{fig:contextual_association}. Note that the contextual GT sampling can be applied with LiDAR-based segmentation maps where we collect top-$k$ closest points and their semantic labels followed by a $k$-NN classifier.

\section{Experiments}
\subsection{Setup}
\myparagraph{Implementation Details.}
Our model is built upon PointPillars~\cite{lang2019pointpillars} architecture, and we follow its default setting for training our model. Our implementation is based on an open source project for LiDAR-based 3D object detection called OpenPCDet~\cite{openpcdet2020}, which supports multiple LiDAR-based 3D perception models, including PointPillars~\cite{lang2019pointpillars}, PV-RCNN~\cite{shi2020pv}, and Voxel RCNN~\cite{deng2021voxel}. Thus, we believe that our proposed regularizing components could be easily applied to other LiDAR-based perception models and ensure reproduction. Our model is trained end-to-end using Adam optimizer with the learning rate 0.003. The whole model is trained for 80 epochs on 4 NVIDIA GeForce RTX 3090 GPUs. Since our data augmentation strategy and gradient balancing technique are turned off during inference, the inference time would remain the same or smaller than that of PointPillars (due to utilizing multi-headed architecture). 

\myparagraph{Evaluation Details.}
For evaluation, we use the widely-used publicly available KITTI~\cite{geiger2013vision} 3D Object Detection dataset, which provides 7,481 training images and 7,518 test images along with LiDAR point clouds. Overall, 80,256 objects are labeled, and (as typically chosen) we focus on three types of objects: cars, pedestrians, and cyclists. Note that our model is based only on LiDAR point clouds during inference, and we use images for better qualitative analysis. Further, we use a large-scale dataset called nuScenes~\cite{caesar2020nuscenes}, which provides over 1500 hours of driving data collected from four different major cities. Our model evaluation is done on ten classes: i.e., car, truck, construction vehicle (CV), bus, trailer, barrier, motorcycle, bicycle, pedestrian, and traffic cone (TC).

\subsection{Quantitative Analysis}

{
 \setlength{\tabcolsep}{4pt}
 \renewcommand{\arraystretch}{1.3} 
\begin{table*}[t]
	\begin{center}
	\caption{3D object detection performance (in mAP) on the publicly available KITTI~\cite{geiger2013vision} validation dataset. We also report counts (in \%) for each class to determine the data imbalance amount. In higher IoU threshold setting: Car (0.7), Pedestrian (0.5), and Cyclist (0.5), In lower IoU threshold setting: Car (0.5), Pedestrian (0.25), and Cyclist (0.25).}
	\label{tab:kitti}
    	\resizebox{\linewidth}{!}{%
    	\begin{tabular}{@{}lcccccccccccccccc@{}} \toprule
    	\multirow{4}{*}{Model} & \multicolumn{8}{c}{Higher IoU threshold setting} & \multicolumn{8}{c}{Lower IoU threshold setting}\\\cmidrule{2-17}
    	& \multicolumn{2}{c}{Car} & \multicolumn{2}{c}{Pedestrian} & \multicolumn{2}{c}{Cyclist} & \multicolumn{2}{c}{Avg.} & \multicolumn{2}{c}{Car} & \multicolumn{2}{c}{Pedestrian} & \multicolumn{2}{c}{Cyclist)} & \multicolumn{2}{c}{Avg.} \\
    	& \multicolumn{2}{c}{(83.00\%)} & \multicolumn{2}{c}{(12.76\%)} & \multicolumn{2}{c}{(4.24\%)} & \multicolumn{2}{c}{(100.00\%)} & \multicolumn{2}{c}{(83.00\%)} & \multicolumn{2}{c}{(12.76\%)} & \multicolumn{2}{c}{(4.24\%)} & \multicolumn{2}{c}{(100.00\%)}\\\cmidrule{2-17}
    	& 3D$\uparrow$ & BEV$\uparrow$ & 3D$\uparrow$ & BEV$\uparrow$ & 3D$\uparrow$ & BEV$\uparrow$ & 3D$\uparrow$ & BEV$\uparrow$ & 3D$\uparrow$ & BEV$\uparrow$ & 3D$\uparrow$ & BEV$\uparrow$ & 3D$\uparrow$ & BEV$\uparrow$ & 3D$\uparrow$ & BEV$\uparrow$ \\\midrule
    	A. PointPillars~\cite{lang2019pointpillars} & 78.04 & 87.49 & 49.40 & 55.78 & 63.95 & 68.97 & 63.80 & 70.75 & {\bf 95.62} & 94.49 & 69.68 & 69.86 & 73.78 & 73.78 & 79.69 & 79.38\\
    	B. A + CBGS~\cite{zhu2019class} & 77.78 & 87.47 & {\bf 51.06} & {\bf 57.40} & 65.53 & 69.40 & 64.79 & 71.42 & 94.33 & 94.51 & 72.20 & 72.57 & 72.54 & 72.56 & 79.69 & 79.88\\\midrule
    	C. A + Ours & 78.37 & {\bf 89.28} & {\bf 51.06} & 56.92 & {\bf 68.79} & {\bf 72.44} & {\bf 66.07} & {\bf 72.88} & 94.88 & {\bf 96.29} & {\bf 73.24} & {\bf 73.36} & {\bf 76.58} & {\bf 76.58} & {\bf 81.57} & {\bf 82.08}\\
    	D. C w/o contextual GT sampling & {\bf 78.77} & 88.28 & 50.50 & 55.86 & 65.31 & 69.24 & 64.86 & 71.13 & 94.42 & 94.61 & 70.64 & 70.88 & 72.45 & 72.81 & 79.17 & 79.43\\
    	\bottomrule
        \end{tabular}}
     \end{center} \vspace{-1em}
\end{table*}
}

\myparagraph{Evaluation with KITTI Dataset.}
We first analyze our proposed model with the publicly available KITTI~\cite{geiger2013vision} 3D object detection dataset. As shown in Table~\ref{tab:kitti}, starting from the baseline (we use PointPillars~\cite{lang2019pointpillars}), we compare the 3D object detection performance (in mAP) with and without our three main components: (i) multiple detection heads per class along with gradient balancing techniques, (ii) pyramid data augmentation, and (iii) contextual GT sampling. We compare our model with CBGS (cross-balanced grouping and sampling, \cite{zhu2019class}), which similarly aims to reduce the negative effect of class imbalance problems in the perception task. 

We observe in Table~\ref{tab:kitti} that our model generally provides a performance improvement in all classes and metrics and generally outperforms the alternative approach. Such an improvement is significant in detecting cyclists, which rarely appear in the dataset (734 out of 17,298 labeled training objects). This may confirm that our proposed components are effective in improving the detection performance of rarely observed objects in such an imbalanced dataset. We also observe that our gradient-balanced multi-headed architecture and contextual GT sampling significantly improves the overall object detection performance, especially for rarely-observed object classes (compare cyclist vs. car and pedestrians)

{
 \setlength{\tabcolsep}{4pt}
 \renewcommand{\arraystretch}{1.3} 
\begin{table*}[t]
	\begin{center}
	\caption{3D object detection performance (in mAP) on the nuScenes~\cite{fong2021panoptic} validation set. We also report counts (in \%) for each class to determine the data imbalance amount. {\it Abbr.} C.V.: Construction Vehicle, Ped: Pedestrian, Moto: Motorcycle, T.C.: Traffic Cone.}
	\label{tab:nuscenes}
    	\resizebox{\linewidth}{!}{%
    	\begin{tabular}{@{}lccccccccccc@{}} \toprule
    	 \multirow{2}{*}{Model} & Car & Ped & Barrier & Truck & T.C. & Trailer & Bus & C.V. & Motor & Bicycle & Avg. \\ \cmidrule{2-12}
    	 & (42.64\%) & (20.31\%) & (13.49\%) & (8.19\%) & (7.90\%) & (2.41\%) & (1.54\%) & (1.39\%) & (1.11\%) & (1.03\%) & (100.00\%) \\ \midrule
    	 A. PointPillars + CBGS & 80.8 & 71.9 & 47.8 & 49.2 & 46.9 & 34.2 & 62.4 & 12.1 & 30.9 & 4.8 & 44.1\\\midrule
    	 B. A + Ours & {\bf 82.1} & 71.9 & \textbf{54.5} & {\bf 53.8} & {\bf 50.1} & {\bf 39.1} & {\bf 67.0} & {\bf 16.3} & {\bf 40.2} & {\bf 9.4} & {\bf 48.4}\\ 
    	 & (11.6$\uparrow$) & (12.0$\uparrow$) & (20.5$\uparrow$) & (28.8$\uparrow$) & (21.7$\uparrow$) & (22.4$\uparrow$) & (32.7$\uparrow$) & (11.8$\uparrow$) & (20.2$\uparrow$) & (7.8$\uparrow$) & (18.9$\uparrow$)\\
    	 C. B + w/o contextual GT sampling & 81.0 & \textbf{72.3} & 50.2 & 49.0 & 45.2 & 34.3 & 63.4 & 10.7 & 32.9 & 6.9 & 44.6\\\bottomrule
        \end{tabular}}
     \end{center} \vspace{-1em}
\end{table*}
}

\myparagraph{Evaluation with a large-scale nuScenes Dataset.}
To further demonstrate the effectiveness of our proposed approach, we evaluate with a large-scale nuScenes dataset, which is more challenging for perception mainly due to its volume and data imbalances across different classes. In Table~\ref{tab:nuscenes}, we compare 3D object detection performance (in 3D and BEV mAPs) for all ten different object classes: (in the sorted order by their numbers) car, truck, bus, trailer, construction vehicle, pedestrian, motorcycle, bicycle, traffic cone, and barrier. Similarly, we compare our model with CBGS~\cite{zhu2019class}. As we observe in Table~\ref{tab:nuscenes}, our model generally outperforms alternatives, and the performance boost is notable in minor classes, such as buses, trailers, construction vehicles, etc. This further confirms that our proposed model is effective in dealing with rarely-observable objects, and such improvement is larger than the existing approach, CBGS~\cite{zhu2019class}. 

\subsection{Ablation Study}

{
 \setlength{\tabcolsep}{4pt}
 \renewcommand{\arraystretch}{1.3} 
\begin{table*}[t]
	\begin{center}
	\caption{The effect of contextual sampling on the 3D object detection performance (in mAP) with three existing object detection models: PointPillars~\cite{lang2019pointpillars}, PV-RCNN~\cite{shi2020pv}, and Voxel RCNN~\cite{deng2021voxel}. We use the publicly available KITTI~\cite{geiger2013vision} validation set.}
	\label{tab:effect_of_contextual_gt_sample}
    	\resizebox{\linewidth}{!}{%
    	\begin{tabular}{@{}lcccccccc@{}} \toprule
    	\multirow{2}{*}{Model} & \multicolumn{2}{c}{Car (0.5)} & \multicolumn{2}{c}{Pedestrian (0.25)} & \multicolumn{2}{c}{Cyclist (0.25)} & \multicolumn{2}{c}{Avg.} \\\cmidrule{2-9}
    	& 3D$\uparrow$ & BEV$\uparrow$ & 3D$\uparrow$ & BEV$\uparrow$ & 3D$\uparrow$ & BEV$\uparrow$ & 3D$\uparrow$ & BEV$\uparrow$ \\\midrule
    	PointPillars~\cite{lang2019pointpillars} + GT Sampling & 94.50 & 94.66 & 70.40 & 70.72 & 71.03 & 71.03 & 78.64 & 78.80\\
    	PointPillars~\cite{lang2019pointpillars} + {\it Contextual} GT Sampling & 94.99 (0.45$\uparrow$) & 95.09 (0.40$\uparrow$) & 70.93 (0.53$\uparrow$) & 71.23 (0.51$\uparrow$) & 72.80 (1.77$\uparrow$) & 72.80 (1.77$\uparrow$) & 79.57 (0.93$\uparrow$) & 79.71 (0.91$\uparrow$) \\\midrule
    	PV-RCNN~\cite{shi2020pv} + GT Sampling & 94.42 & 96.20 & 75.32 & 75.60 & 79.15 & 79.15 & 82.96 & 83.65\\
    	PV-RCNN~\cite{shi2020pv} + {\it Contextual} GT Sampling & 94.74 (0.32$\uparrow$) & 96.50 (0.30$\uparrow$) & 75.50 (0.18$\uparrow$) & 75.92 (0.32$\uparrow$) & 83.25 (4.10$\uparrow$) & 83.25 (4.10$\uparrow$) & 84.50 (1.54$\uparrow$) & 85.22 (1.57$\uparrow$)\\\midrule
    	Voxel RCNN~\cite{deng2021voxel} + GT Sampling & 94.91 & 96.66 & - & - & - & - & 94.91 & 96.66 \\
    	Voxel RCNN~\cite{deng2021voxel} + {\it Contextual} GT Sampling & 97.08 (2.17$\uparrow$) & 97.31 (0.65$\uparrow$) & - & - & - & - & 97.08 (2.17$\uparrow$) & 97.31 (0.65$\uparrow$)\\
    	\bottomrule
        \end{tabular}}
     \end{center} \vspace{-1em}
\end{table*}
}

\begin{figure}[t]
    \begin{center}
        \includegraphics[width=\linewidth]{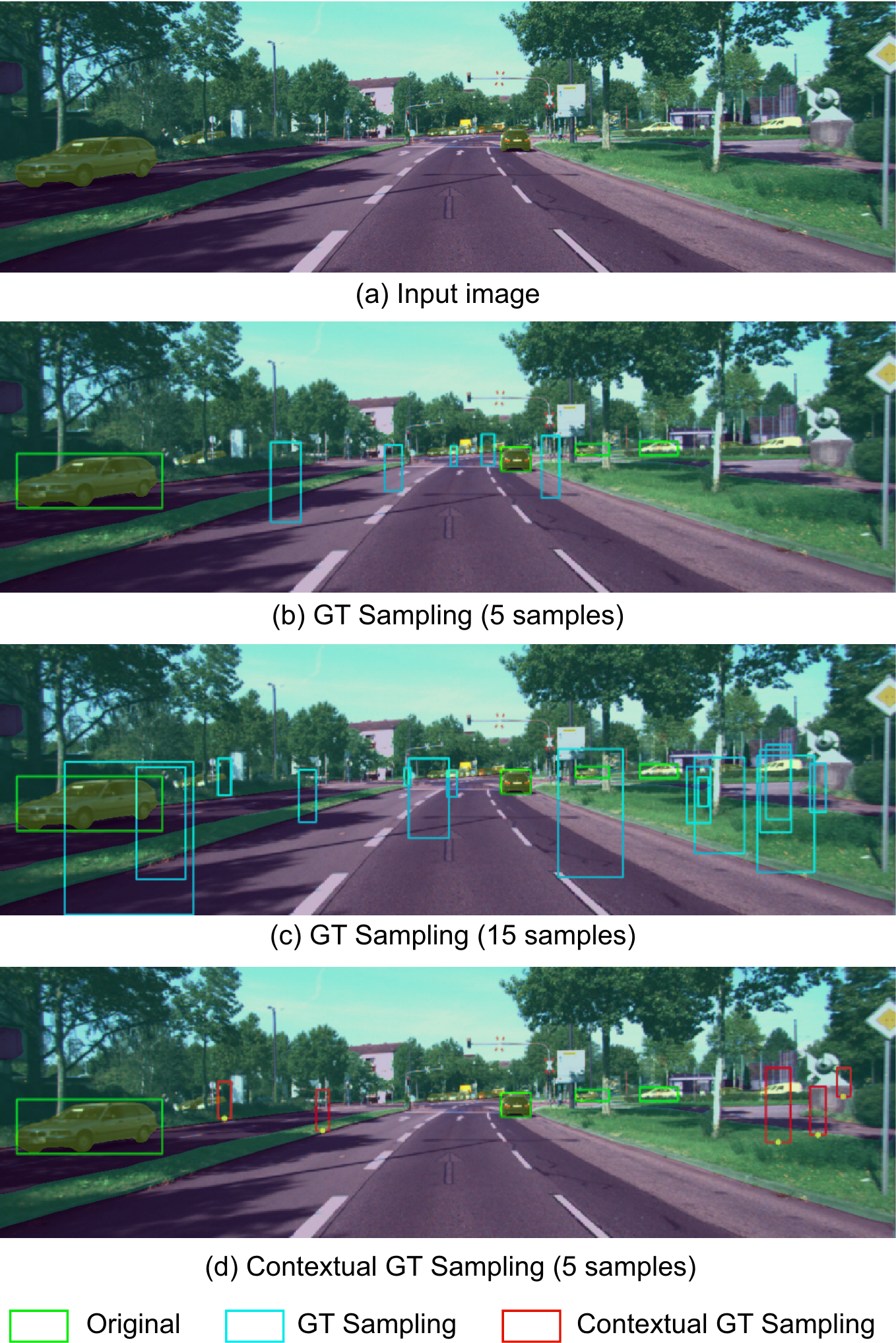}
    \end{center}
    \caption{A comparison between existing ground-truth (GT) sampling and our proposed contextual ground-truth (GT) sampling. (a) A front-view image overlaid by a semantic segmentation map. An image with original bounding boxes (green) and augmented bounding boxes (cyan) from a ground-truth bounding box database either by (b-c) GT sampling method or (d) our proposed contextual GT sampling method.}
    \label{fig:contextual_sampling}
\end{figure}

\myparagraph{Effect of Contextual GT Sampling for LiDAR Data Augmentation.}
We further evaluate the effect of using the contextual sampling with existing LiDAR-based 3D object detection models: PointPillars~\cite{lang2019pointpillars}, PV-RCNN~\cite{shi2020pv}, and Voxel RCNN~\cite{deng2021voxel}. As we observe in Table~\ref{tab:effect_of_contextual_gt_sample}, perception performance (in terms of 3D mAP and BEV mAP) generally improves by changing conventional GT sampling with our contextual GT sampling. A similar pattern of improvements is observed in all models, and a larger effect is obtained in minor classes, such as pedestrians and cyclists. Note that we only consider cars for Voxel RCNN~\cite{deng2021voxel} as its original architecture is focusing only on detecting cars.

In Figure~\ref{fig:contextual_sampling}, we provide an example of conventional GT sampling and our proposed contextual GT sampling for pedestrians. A probability occupancy grid is computed for augmented LiDAR points to be copied (from a GT LiDAR point cloud database) and pasted (into a scene), given the semantic information. For example, ground-truth pedestrian points are sampled from a database and placed into the scene based on the probability occupancy grid (compare how points are augmented by GT sampling (cyan) and our contextual GT sampling (red)). Note that a differently color-coded semantic segmentation map overlays all images.

{
 \setlength{\tabcolsep}{4pt}
 \renewcommand{\arraystretch}{1.3} 
\begin{table}[t]
	\begin{center}
	\caption{We compare 3D object detection accuracy with three different multitask learning techniques: GradCosine~\cite{du2018adapting}, GradNorm~\cite{chen2018gradnorm}, and Dynamic Weight Average DWA~\cite{liu2019end}. We report scores on the KITTI~\cite{geiger2013vision} validation set with a set of higher IoU threshold.}
	\label{tab:effect_of_grad_bal}
    	\resizebox{.9\linewidth}{!}{%
    	\begin{tabular}{@{}lcc@{}} \toprule
    	Model & 3D$\uparrow$ & BEV$\uparrow$ \\\midrule
    	A. PointPillars~\cite{lang2019pointpillars} & 63.80 & 70.75 \\
    	B. A + Per-Class Multiple Detection Heads & 64.07 & 71.03 \\ \midrule
    	C. B + GradCosine~\cite{du2018adapting} & 63.60 & 71.09 \\
    	D. B + GranNorm~\cite{chen2018gradnorm} & 63.65 & 69.72\\
    	E. B + DWA~\cite{liu2019end} & {\bf 64.86} & {\bf 71.13} \\
    	\bottomrule
        \end{tabular}}
     \end{center} \vspace{-1em}
\end{table}
}

\myparagraph{Effect of Gradient Balancing.}
In Table~\ref{tab:effect_of_grad_bal}, we further provide our ablation study to verify the effect of balancing gradients across multiple per-class detection heads. Given the PointPillars~\cite{lang2019pointpillars} as a baseline, we first modify the network architecture with per-class multiple detection heads (Model B). Then, we apply the following three multitask learning techniques: GradCosine~\cite{du2018adapting}, GradNorm~\cite{chen2018gradnorm}, and Dynamic Weight Averaging (DWA, \cite{liu2019end}). We observe in Table~\ref{tab:effect_of_grad_bal} that (i) applying per-class multiple detection heads generally improves the overall detection accuracy. Also, we observe that (ii) using Dynamic Weight Averaging outperforms other gradient balancing techniques, and DWA provides a performance gain potentially due to the balanced losses across different heads. Interestingly, the other two techniques (GradCosine and GradNorm) generally degrade the overall detection performance. This is probably due to a data imbalance problem and indicates that a contextual GT sampling technique needs to be used together.

\begin{figure*}[t]
    \begin{center}
        \includegraphics[width=\linewidth]{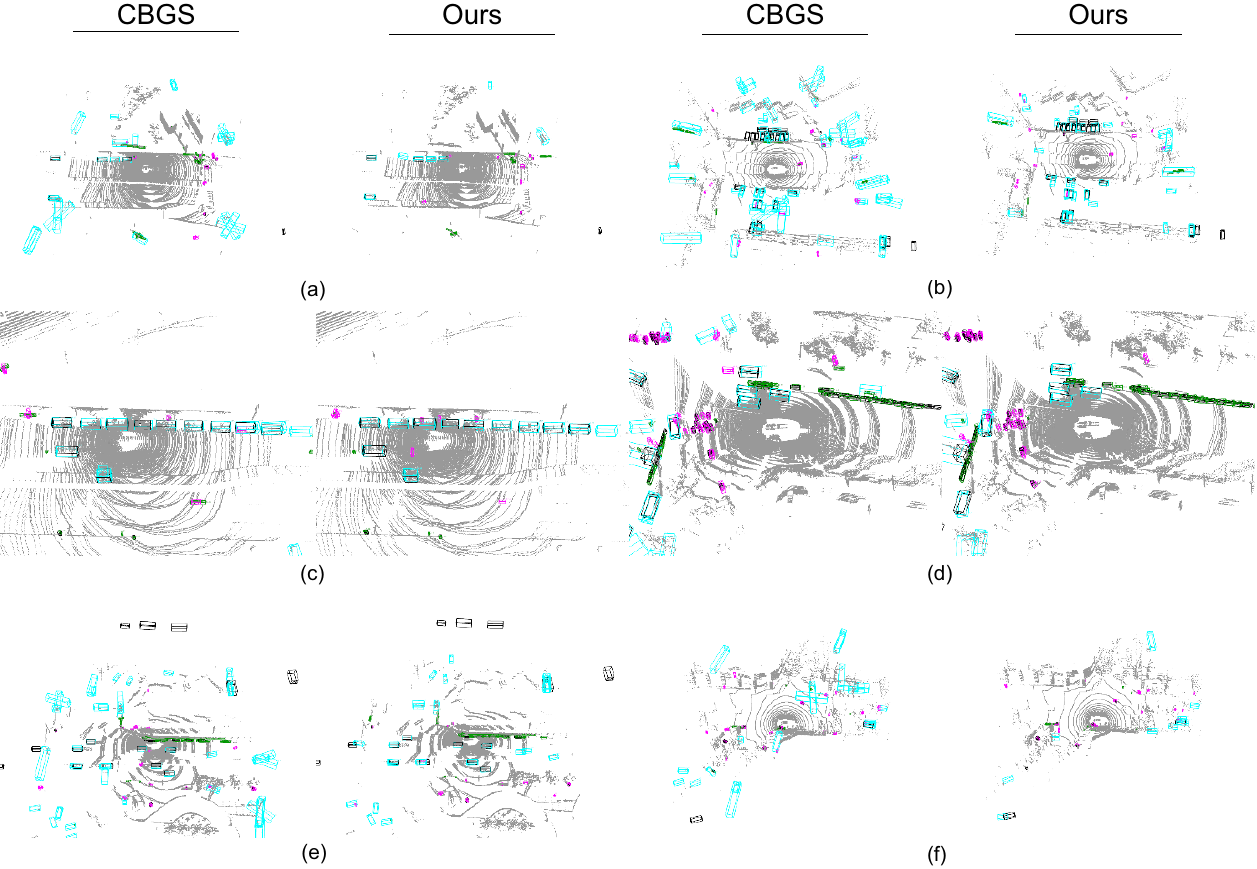}
    \end{center}
    \caption{Comparison between CBGS (based on PointPillars~\cite{lang2019pointpillars}) vs. Ours (PointPillars + Per-Class Multiple Detection Heads with Gradient Balancing + Contextual GT Sampling) on nuScenes validation set. Ground truth boxes are color-coded as black, while other predicted boxes are color-coded as cyan (Car, Truck, Construction Vehicle, Bus, Trailer), pink (Pedestrian, Bicycle, Motocycle), and green (Barrier, Traffic Cone). } 
    \label{fig:qualitative_analysis}
\end{figure*}

\subsection{Qualitative Analysis}
\myparagraph{Analysis with nuScenes dataset.}
In Figure~\ref{fig:qualitative_analysis} (a-f), we provide a qualitative comparison of predictions between our baseline (PointPillars+CBGS) and ours (PointPillars with our proposed per-class multi-heads with gradient balancing and contextual GT sampling). We provide six examples sampled from the nuScenes validation dataset with different color-coded bounding boxes (see caption). We observe that ours generally predict fewer false positives, especially for minor classes (see cyan boxes). This is probably because our per-class multiple detection heads encourage the model to learn more class-specific features, resulting in better robustness in detection. Further, we also observe that our model predicts objects in a more reasonable location. As we use contextual GT sampling, which considers more realistic places to augment objects, we observe that ours generally predicts objects in a more right place. For example, our baseline model produces prediction outputs of trucks in some counter-intuitive places. 

\begin{figure}[t]
    \begin{center}
        \includegraphics[width=\linewidth]{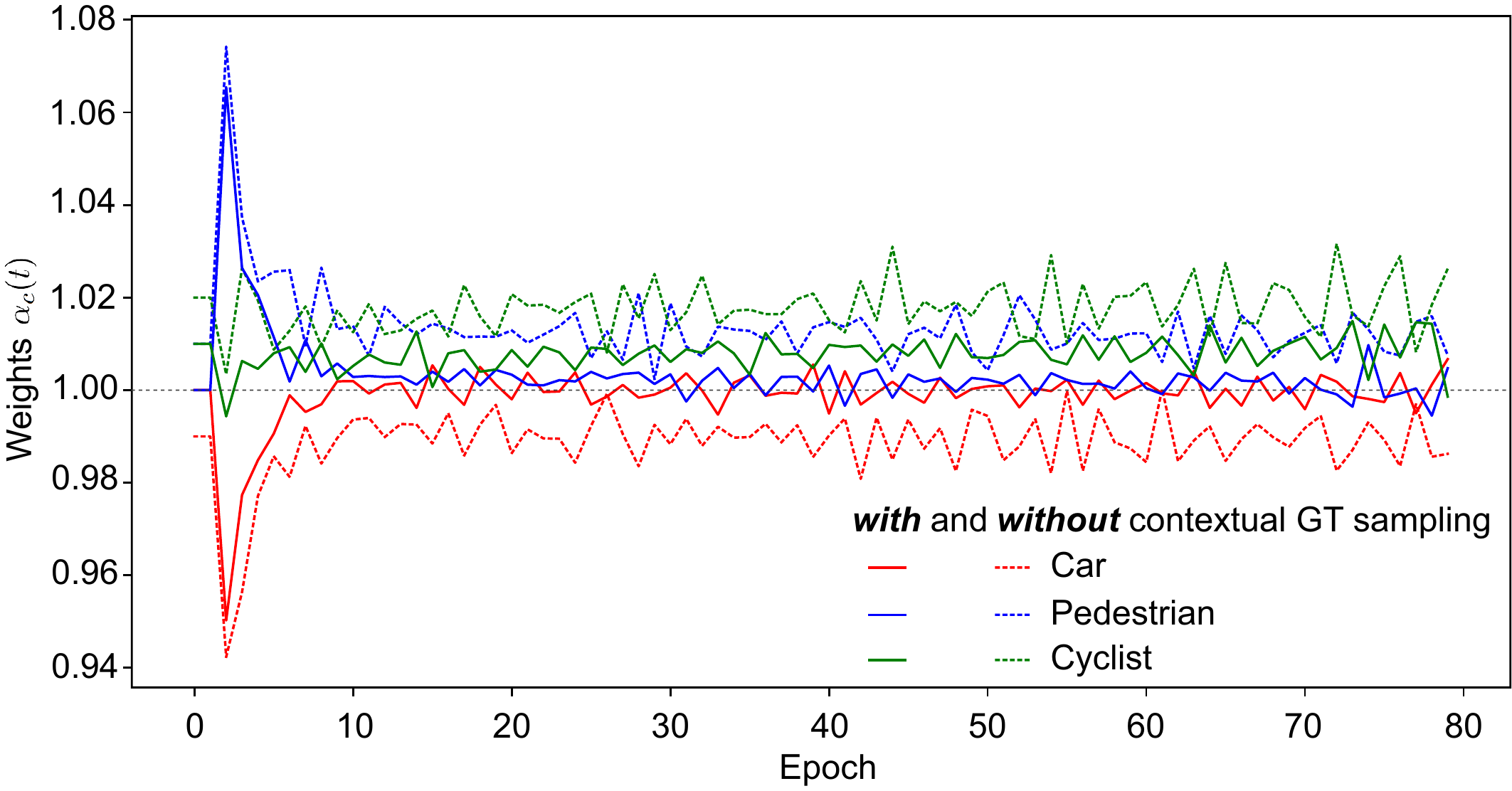}
    \end{center}
    \caption{Changes in balancing weights $\alpha_{c}(t)$ (measured at the end of each epoch) for different classes: cars (red), pedestrians (blue), and cyclists (green). We also compare weights with (solid) and without (dashed) contextual GT sampling. Data: KITTI.}
    \label{fig:dwa}
\end{figure}

\myparagraph{Weight Changes by Dynamic Weight Average.}
We also analyze weight ($\alpha_{c}(t)$) changes that balance losses from each detection head to reduce the data imbalance problem. Without contextual GT sampling (dashed), we observe that the model provides more weights on minor classes (compare green (cyclists) vs. red (cars)) to balance between two heads. This trend continues even with contextual GT sampling, but their weight gap is reduced. This is because our contextual GT sampling provides more examples for minor classes to balance the number of examples across classes.

\myparagraph{Social Impact.}
We believe our effort to reduce the data imbalance problem is also in the mainstream of ethical AI, which focuses on removing implicit biases potentially from training data that might include biased data collection or reflect historical or social inequalities. We aim to make the model pay more attention to unrepresentative classes, resulting in lower error rates for minor classes while retaining the same or lower error rates for others. Also, as our model is for building autonomous driving systems, our work would inherit its social impact.

\section{Conclusion}
In this work, we introduced a method to address the data imbalance problem in the LiDAR-based 3D object detection task. We proposed two main components: (1) multi-task learning-inspired per-class multi-headed LiDAR-based 3D object detector where losses from each head are modified to be balanced. (2) Contextual ground truth (GT) sampling, which improves the conventional GT sampling by leveraging semantic scene information to introduce objects in a more realistic location, resulting in a better quality of data augmentation. We conducted various experiments with a large-scale nuScenes dataset and a widely-used KITTI dataset. We demonstrated the effectiveness of our proposed method by showing improved accuracies of minor classes. 

\myparagraph{Acknowledgements.}
This work was supported by the grant from Autonomous Driving Center at Hyundai Motor Company's R\&D Division.

{\small
\bibliographystyle{ieee_fullname}
\bibliography{egbib}
}

\end{document}